\documentclass[]{article} 

\usepackage{url,hyperref,lineno,microtype,algorithmic,graphicx,amsmath,amssymb}
\usepackage[utf8]{inputenc}
\usepackage[margin=1in]{geometry}
\usepackage[sort&compress,numbers,square]{natbib}
\title{Pruning coupled with learning, ensembles of minimal neural networks, and future of XAI}

\author{Alexander N Gorban\,$^{1,2}$ and Evgeny M Mirkes\,$^{1,2}$}
\date{
$^{1}$Department of Mathematics, University of Leicester, Leicester, UK \\ \texttt{\{ag153, em322\}@le.ac.uk}\\
$^{2}$Laboratory of Advanced Methods for High-dimensional Data Analysis, \\ 
Lobachevsky State University, Nizhny Novgorod, Russia}

\begin{document}

\maketitle

\begin{abstract}
Pruning coupled with learning aims to optimize the neural network (NN) structure for solving specific problems. This optimization can be used for various purposes: to prevent overfitting, to save resources for implementation and training, to provide explainability of the trained NN, and many others. The minimal structure that cannot be pruned further is not unique. Ensemble of minimal structures can be used as a committee of intellectual agents that solves problems by voting. Each minimal NN presents an ``empirical knowledge" about the problem and can be verbalized. The non-uniqueness of such knowledge extracted from data is an important property of data-driven Artificial Intelligence (AI).  In this work, we review an approach to pruning based on the principle: {\it What controls training should control pruning.} This principle is expected to work both for artificial NN and for selection and modification of important synaptic contacts in brain. In back-propagation artificial NN learning is controlled by the gradient of loss functions. Therefore, the first order sensitivity indicators are used for pruning and the algorithms based on these indicators are reviewed. The notion of logically transparent NN was introduced. The approach was illustrated on the problem of political forecasting: predicting the results of the US presidential election. Eight minimal NN were produced that give different forecasting algorithms.  The non-uniqueness of solution  can be utilised by creation of expert panels (committee). Another use of NN pluralism is to identify areas of input signals where further data collection is most useful. In conclusion, we discuss the possible future of widely advertised XAI program.
\end{abstract}

\section{Introduction}

Artificial neural networks (NN) with an arbitrary activation function are universal approximators for continuous functions and maps \cite{Hornik1991, Kreinovich_1991, Gorban_1998}. This is a mathematical basis for use of NN in many applications. However, this universality creates another problem: what is required NN structure? How many neurons and connections are needed to solve a particular problem? Is it possible to simplify a particular NN without loss of performance?  

\begin{figure*}[ht!]
\centering
\textbf{(A)}\includegraphics[width=0.45\textwidth ]{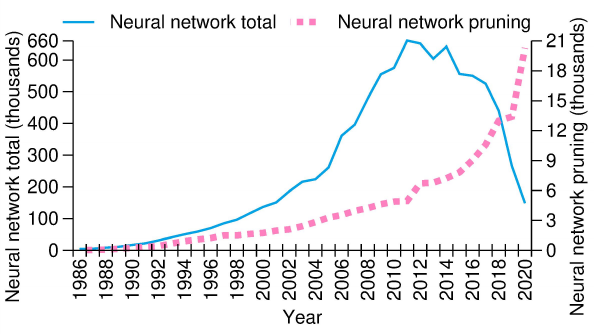}\;\;
\textbf{(B)}\includegraphics[width=0.45\textwidth ]{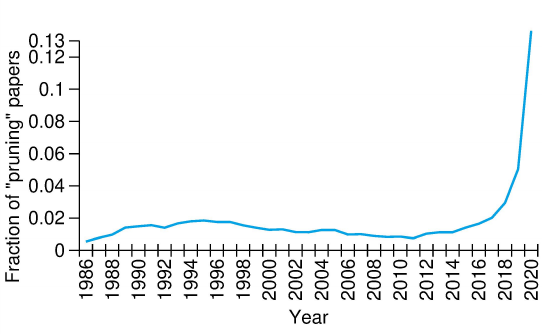}
\caption{Research documents according to Google Scholar (found in June 2021) for different years of publication: (A) Number of documents for the search queriy  \textbf{\emph{``neural'' AND ``network''  AND ``pruning''}} (dotted pink line), and for the search query \textbf{\emph{``neural'' AND ``network''}} (solid blue line), and (B) ratio of these numbers.}
\label{fig:Develops}
\end{figure*}

The trend of publications related to NN found by Google Scholar is presented in Figure~\ref{fig:Develops}A. According to these data, the intensity of publications about NN  is currently decreasing, whereas publications about NN pruning are growing exponentially. The fraction of NN publications which mention pruning demonstrates fast explosion (Figure~\ref{fig:Develops}B).

The question ``Neural net pruning - why and how" \cite{Sietsma_1988}  was asked very soon after the seminal back-propagation paper published by \cite{RUMELHART1986}. Already at 1990, a system of methods for NN pruning was proposed \cite{Sietsma_1988, mozer1989skeletonization, Janowsky_1989, Karnin, Whitley_1990, Hagiwara_1990, gorban1990training, lecun1990optimal}.

Usually, pruning is defined as removing synapses from NN with acceptable loss of performance (or even without such a loss). There could be special requests for pruning like removing whole nodes (neurons) or even bigger blocks and reducing number of layers \cite{Gorban_2019}. Most algorithms of pruning include steps of removing, training of pruned network, and testing. If a pruned network cannot achieve the desired performance after training then the pruning is considered to be `too radical' and some of the removed elements return to the NN. Pruning without additional training is called `clipping' \cite{Janowsky_1989}. Simplification of NN by combination of clipping and testing can be also iterative.   

When pruning, the values of the selected weights are reset to zero. A more general simplification assumes `quantization', reduction of the set of possible values of weights to several selected numbers (a sort of numerosity reduction). This transformation is important for implementing trained NN in small gadgets and, in particular, for development smaller and `greener' NN with less energy consumption. The algorithms for quantization are similar to pruning: instead of resetting the weights to zero, they are sent to the nearest selected values \cite{Han_2015}.

Nowadays, thousands of works about NN pruning were published and the interest to this area increases very fast (Fig.~\ref{fig:Develops}). In this paper, we analyse various approaches to NN pruning and extraction of explicit knowledge from NN.

The paper is organised as follows: Section~\ref{GenPrun} describes methods for estimating the sensitivity indicators, Section~\ref{PrunProc} describes the pruning procedures and strategies for various pruning problems, Section~\ref{KnowExtr} describes pruning procedure which are specific for knowledge extraction and presents a example of knowledge extraction for the task of prediction the result of the USA presidential election. 
A simple local statistical approach to pruning based on sparse linear regression for each single neuron is outlined in Section~\ref{sparse}. Section~\ref{discus} presents discussion of proposed methods and the problem of biologically relevant pruning.

\section{Why NN pruning?}\label{sec:Why}

Why NN pruning attracts so much attention during so long time? 
Several groups of reasons were formulated by developers:
\begin{itemize}
\item Improvement the performance of learning algorithms and the resulting NNs to provide faster learning, better generalization abilities, etc.
\item New service to users, first of all, explaining decisions on NN and transforming implicit skills of NN into explicit knowledge.
\item Simplify structure for cheaper hardware implementations of NN.
\end{itemize}

In particular, pruning was considered as a tool for dealing with overfitting \cite{Sietsma_1991}. It is a commonly accepted and well-proved idea in statistics that the number of parameters in the model  should be penalised in the process of the model selection. There are different forms of the trade-off between the model accuracy (likelihood) and the number of parameters in comparison of models, for example, the Schwarz information criterion, the Akaike information criterion and their generalizations \cite{Ding_2018}.

Pruning decreases the number of parameters in fitting and, therefore, is proposed as one of the tools for dealing with overfitting among the other tools, like validation--based early stopping rules \cite{Amari_1997}, data augmentation (reviewed by \cite{Shorten_2019}), random perturbation of weights in the curse of learning (with or without averaging of predictions) \cite{gorban1990training,gorban1996neural}, or very popular now dropouts \cite{Srivastava_2014}.
Special regularization through the loss function can also significantly reduce the effects of overparametrisation. The methods of NN growth \cite{Ishikawa1996, Murase_1991, fahlman1990cascade} aim to grow NN without overparametrization instead of reduction of large NN.  

The expectation that pruning accelerates learning is not always true. If we take a NN with minimized structure and good performance and start to train it by  gradient descent with  new initialization of weigh, the training process may be pathologically slow or even unsuccessful \cite{gorban1990training} (of course, this result may depend on the details of the learning algorithm).

Recent observations show that in deep learning large number of parameters does not cause overfitting, and overparametrized NN have outstanding generalization power, at least, in some cases \cite{berner2021}. Therefore, two strategies are possible to prevent overfitting: (i) reduce the number of parameters and (ii) increase the number of parameters. This seems paradoxical, but it is in line with current knowledge. This is a serious challenge to the mathematical theory of learning.

The second NN drawback is unreadability: trained NN solves the problem, but we cannot understand and explain this solution \cite{Elizondo2005}. Explainable AI is needed \cite{Gunning_2019}. This problem can also be solved by the training-pruning process (see, for example, the early works by \cite{gorban1996neural, mirkes1998neurocomputer, gordienko1993construction, Ishibuchi} or recent versions by \cite{LiuWu_2019}) with the subsequent interpretation of NN structure. Detailed description of this three--step approach is presented in \cite{Gorban1999Generation} and in this paper. 

A result of this pruning is one or more algorithms described in natural language -- and explained AI decision. Very often, the explanation is fundamentally non-unique unless we artificially restrict the set of possibilities. The presence of several different algorithms may be considered as drawback of the method but, from the other side, several algorithms can be used to diagnose the situations, where the solutions are doubtful (because the decisions of different algorithms do not coincide). The areas, where the outputs of pruned systems are significantly different can be recommended for further data collection. Thus, the multiplicity of explanations provides additional opportunities for analytical work. Below,  we demonstrate such non-uniqueness for NN analysis of USA presidential elections.

The third problem is the implementation of trained NNs. It is almost impossible to use large NN and extended or double precision variables for small gadgets (like mobile phones or Raspberry Pi) NN implementation. Such a product will work slow and require a lot of memory.  Moreover, it is not necessary because of redundancy of NN. Pruning simplifies NN,  reduces required precision and creates more efficient implementations  \cite{Gorban1999Generation, Keegstra, molchanov2017pruning}. This approach also has an alternative implementation in the form of growing NN \cite{Hoehfeld_1992}. Another problem that is closely related to this is the problem of the robustness of NN with respect to input noise. Reduction of implementation precision increases robustness.

The fourth problem is closely related to the third and can be formulated as the problem of backward feature selection \cite{kira1992feature}: select the minimal set of input features that provides a solution to the problem with the required accuracy. Back propagation NN provides the ability to calculate derivative of loss function with respect any elements of NN and with respect to any output, intermediate, and input signals. This allows the user to identify the influence of each input feature and remove one or more of the least important.

All the  problems described above were formulated already for the so-called ``shallow'' NN at  the first NN pruning peak in 1990-2000 (see Figure~\ref{fig:Develops}). The new generation of NN techniques, Deep Learning (DL) NN and Convolutional Neural Network (CNN) began to be widely used in the early 21st century (see review by \cite{SCHMIDHUBER201585}). The number of papers related to DL NN pruning monotonously increased during this period (see, for example, works by \cite{Han_2015, han2015learning, He_2017, Tung_2018, Gorban_2019}. The robustness of DL NN solutions is one of the main problems in the DL NN applications. The study of DL NN stability is presented by \cite{Su_2019, Mani, madry2017towards, Deng_2014}.

As was suggested by \cite{Fu_Jie_Huang}, it is possible to use trained DL CNN without final classification layers (usually one or two fully connected layers) as a feature generator, and then apply any classification methods in this space of generated features. This approach was also used for person re-identification problem \cite{Taigman_2014, Sun_2015}. \cite{Gorban_2019} describe  the use of trained CNN as a generator of features with different number of layers in the generating part. It is shown that reducing the depth of the generating part may increase  the number of errors but, nevertheless, allows to solve problem with high enough accuracy.

We can conclude that NN pruning can be used not just for removing of redundant weights but for solving of many related problems:
\begin{enumerate}
  \item Feature selection: removal of neurons in the input layer. 
  \item Identification the appropriate NN architecture to solve the problem and prevent overfitting. For this, various NN elements are removed (usually neurons in shallow NN or filters in CNN).
  \item  Reduction of the precision of synaptic weights to provide cheap and fast implementation of trained NN.
  \item Replacement of the activation function of neurons with a simple function. For example, a threshold or piecewise linear function instead of a sigmoid one. Solving this problem also reduces the cost of implementing NN.
  \item Uniform network simplification with a decrease in the maximum number of synapses associated with each neuron. This pruning problem was introduced especially for the problem of knowledge extraction: it is easier to verbalise the functioning of a neuron  if it has a small number of input signals.
  \item  Pruning by removing the last layers of deep CNN and usage of CNN to generate features 
\end{enumerate}

Removing of redundant synapses may be formulated in the framework of the general sparse regression problem. For example,  LASSO penalties for the values of coefficients can be used  \cite{Tibshirani_1996}. Let performance of the NN be evaluated by the loss function $L$. Back propagation learning includes gradient or stochastic gradient minimization of $L$ on the training set using various preprocessing procedures, augmentation of learning algorithms, and stopping criteria. Sparsity of the NN can be achieved by the additional penalty terms (in the  Lagrangian form): minimization of
$$L+\lambda \sum_i |w_i|^p,$$  
where $|w_i|$ are weights, $\lambda>0$, and $1\geq p >0$ (the classical form uses $p=1$).

When $|w_i|$ becomes small enough, the weight is removed from the NN.

For NN surgery with various structural requirements, the methods of sensitivity indicators is used more often. There are several approaches to constructing sensitivity indicators. The simplest idea is: smaller weights are less important and should be removed first. The objections are obvious: this criterion does not take into account the signals, but the weights are coupled with the signals in the NN functioning, and the products 
weight$\times$signal are transmitted further in the in the networks. Importance of synapses should depend on the data and the internal signals produced by the NN. Nevertheless, this approach, surprisingly, works, and  pruning the low-weight connections may give encouraging results \cite{han2015learning, Zhao_2021}. For more precise evaluation of parameters importance, the second-order sensitivity indicators are proposed  based on the second-order Taylor expansion of $L$ near optimum \cite{hassibi1993second, Hassibi, lecun1990optimal}. But computation of $L$ Hessian requires large additional resources. The first-order sensitivity indicators based on the first-order Taylor approximations of $L$ takes into account the values of inputs and internal signals, and utilises the same operation of back propagation for computation of $L$ gradients. 

\section{First order sensitivity indicators}\label{GenPrun}

To prune NN we need to know the importance of each NN element. There are several different measures of this importance and different names for them: `sensitivity indicators' \cite{gorban1990training, gorban1998neuroinformatics, mirkes1998neurocomputer}, `effectiveness of hidden units' \cite{Majima_1994, yoshimura1992new}, `second order sensitivity analysis' \cite{gorban1990training, lecun1990optimal, Hassibi, hassibi1993second}.  \cite{gorban1990training, gorban1996neural} described many different approaches, but recommended using first-order sensitivity indicators that combine acceptable accuracy with computational efficiency.

\begin{figure}[ht!]
\centering{
\includegraphics[width=0.3\columnwidth ]{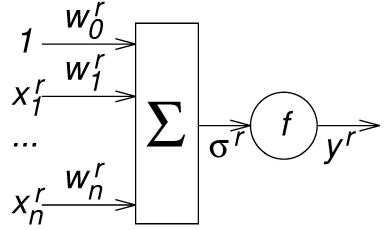}}
\caption{The regular representation of the artificial neuron with number $r$: 1 is a constant unit signal, $x_i$ are input signals of the neuron, $w^r_0$ is the bias, $w^r_i$ are the synaptic weights, $\Sigma$ is the summation element, $\sigma^r$ is its output, $f$ is an activation function, usually non-linear.}
\label{fig:neuron}
\end{figure}

Let us consider simple feedforward NN. The same approach can be used for more complex NN, but the detailed description will be slightly more cumbersome. NN consists of neurons grouped into layers. The standard representation of an artificial neuron is shown in Figure~\ref{fig:neuron}. A standard formal neuron with number $r$ contains linear function 
\begin{equation}\label{combiner}
\sigma^r=w_0^r+\sum_{i=1}^{n}x_iw^r_i,
\end{equation}
where $x_i$ is the $i$th input of the neuron, $w_i^r$ is the $i$th synaptic weight of the neuron, $w_0^r$ is the bias of the neuron). 

The output of the neuron is $y^r=f(\sigma^r)$, where $f$ is the nonlinear activation function $y^r=f(\sigma^r)$. The linear element that calculates the weighted sum (\ref{combiner}) is often called the adaptive linear combiner.

Further we consider pruning based on first-order sensitivity indicators..

NN receives a $d$ dimensional input vector $u$. This input vector propagates through NN, and for each neuron, the output values $y$ are calculated and, finally, the NN output value $\hat{z}$ is calculated. In supervising learning, for the training and test examples, the desired (`proper') output $z$ is known and the loss function $L^j=L(z^j,\hat{z}^j)$ is defined (it can be usual mean square deviation function or more sophisticated loss functions with soft margins  \cite{gorban1990training, gorban1996estimations, mirkes1998neurocomputer}). Further back propagation allows us to calculate the derivatives of the loss function with respect to each input signal of NN and synaptic weight, bias, and neuron output:
$$\frac{\partial L }{\partial u_i}, \frac{\partial L }{\partial w_i^r}, \frac{\partial L}{\partial y^r}.$$
Detailed algorithms of all this derivatives calculation  can be found elsewhere.

Let us have training set with $N$ pairs $(u^j, z^j)$. The inputs, outputs, loss functions and other quantities for the $j$th sample will be marked by the superscript $j$. The goal of the training process is to minimise the total loss function, which is the sum of the individual loss functions:

$$L=\sum_{j=1}^{N}L^j,$$
where $L^j=L(u^j, z^j)$.

The stopping criteria may require to interrupt minimization earlier when the signs of performance decrease because overfitting are detected on a special test set.  
 
Consider the problem of feature selection. For each individual element $(u^j, z^j)$ from the training set, we can apply the Taylor formula for the loss function with respect to input signals

$$L^j(v^j)=L^j(u^j)+\sum_{i=1}^{d}\frac{\partial L^j}{\partial u^j_i}(v^j_i-u^j_i)+o(\|v^j-u^j\|),$$
where $v^j$ is the modified input vector. Since we are interested in feature selection we can consider vector $v^j$ which is the same as $u^j$ excluding one element: $v^j_i=u^j_i(1-\delta_{ik})$, where $\delta_{ik}$ is the Kronecker delta. In this case we can evaluate the cost of removing the input feature $k$ in linear approximation as

\begin{equation}\label{chi}
\chi^j_k=\left|\frac{\partial L^j}{\partial u^j_k}u^j_k\right| (\approx \left|L^j(v^j)-L^j(u^j)\right|).
\end{equation}

The sensitivity indicator is $\chi^j_k$. It is clear that this indicator depends on the element of the training set.   There are several possible methods for combining the gradients $ L $ found for various training inputs into the sensitivity of the NN performance on the entire training set to removal of the input. 

For example, we can define maximal or average indicator:

\begin{equation}
\chi^{max}_k=\max_{j=1,\ldots,N}\chi^j_k,\;\;\;\chi^{avg}_k=\frac{1}{N}\sum_{j=1}^N\chi^j_k.
\label{eq:inp}
\end{equation}

$\chi^{max}_k$ evaluates the maximum influence of the input signal (this signal has never been more important than $\chi^{max}_k$). A widely used alternative is the mean influence or simple sum:  $\chi^{avg}_k$ evaluates the average influence of the input signal.

The components of the batch gradient can be also used as sensitivity indicators: instead averaging the absolute values (\ref{chi}) found for single examples, the sum of the linear approximations can be evaluated first:

\begin{equation}\label{chiB}
\chi^{\rm B}_k=\left|\sum_j\frac{\partial L^j}{\partial u^j_k}u^j_k\right| \left(\approx \left|\sum_j(L^j(v^j)-L^j(u^j))\right| \right).
\end{equation}

For the modification of synaptic weights (include bias) the value of the loss function when $w^r_i$ changes by $v^r_i$ is

$$L^j(v)=L^j(w)+\sum_{r=1}^{m}\sum_{i=0}^{n^r}\frac{\partial L^j}{\partial w^r_i}(v^r_i-w^r_i)+o(\|v^r_i-w^r_i\|),$$

where $m$ is the number of neurons, $n^r$ is the number of input signals of neuron $r$, $v^r_i$ is the new value of weight $i$ of neuron $r$.

The modification includes removal of the weight or change it to the closest (`cheap') selected value.  The set of preferred values can be different for different problems. For example, to extract explicit knowledge from NN it is useful to use weights only with values from the set $S=\{-1, 0, 1\}$. For a low-cost implementation it is possible to convert weights to an integer and apply cheaper software (see, e.g. \cite{Bertin_2004}). For the problem of removing synaptic weights the preferred value is $0$. 

The effect of changing of one weight $w^r_i$ only ($v^k_l=w^k_l\;\forall k\ne r \text{ or } i\ne l)$ in linear approximation is

$$\chi^{js}_k=\left|\frac{\partial L^j}{\partial w^s_k}(w^s_k-v^s_k)\right| \left(\approx \left|L^j(v)-L^j(w)\right|\right).$$
For the problems of removing or reducing of precision of synaptic weights the sensitivity indicators $\chi^{js}_k$ depend on the element of the training set. In this case we can use modification of formulae \eqref{eq:inp} to evaluate the impact for the entire training set:

\begin{equation}
\chi^{max}_{sk}=\max_{j=1,\ldots,N}\chi^{js}_k =\left|w^s_k-v^s_k\right|\max_{j=1,\ldots,N}\left|\frac{\partial L^j}{\partial w^s_k}\right|,
\label{eq:max2}
\end{equation}
\begin{equation}
\chi^{avg}_{sk}=\frac{1}{N}\sum_{j=1}^N\chi^{js}_k=\frac{\left|w^s_k-v^s_k\right|}{N}\sum_{j=1}^N\left|\frac{\partial L^j}{\partial w^s_k}\right|.
\label{eq:avg2}
\end{equation}

The batch sensitivity indicators for weights are defined analogously to (\ref{chi}):
\begin{equation}\label{chiBW}
\chi^{\rm B}_k=\left|(v^s_k-w^s_k)\sum_{j=1}^N \frac{\partial L^j}{\partial w^s_k}\right|.  
\end{equation}

The problem of neuron removing can be solved by removing synaptic weights. This approach is not the best because in the bad case the number of synaptic weights can be drastically reduced without removing any neurons. In the worst case, each neuron may have only one synaptic weight but all neurons will affect NN output. To avoid such situation, we can consider the sensitivity indicators to the entire neuron instead of synaptic weights. To evaluate these indicators we can use the first order Taylor formula again and get

$$\chi^j_k= \left|\frac{\partial L^j}{\partial y^{jk}}y^{jk}\right|,$$
where $y^{jr}$  is the output of the neuron $r$ for the element $j$ of the training set.

To evaluate the sensitivity indicator for the entire training set, one of the formulae \eqref{eq:inp} can be used. The similar approach for neuron removal problem was tested  by \cite{Siegel_2016}.

The distribution of sensitivity indicators   on the training and test sets can be used for many purposes, for example, for evaluation of probability of large performance loss after NN modification.

\begin{figure}[ht!]
\centering
\includegraphics[width=0.49\columnwidth ]{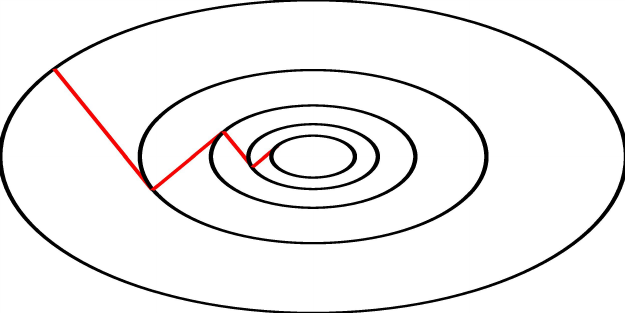}
\caption{Four consecutive steps of steepest descent method for 2D quadratic form}
\label{fig:SDM}
\end{figure}

Now we have sensitivity indicators for the problems of input feature selection, removing or reducing precision of synaptic weights, and neuron removing. All of these indicators are based on the gradient of the loss function. However, gradient descent optimization has one obvious property \cite{KANTOROVICH_1982}: if we precisely find minimum in the direction of the anti-gradient, then the gradient calculated at the new point will be orthogonal to the previous gradient. Several consecutive steps of steepest gradient descent for the simplest 2D quadratic form are presented in Figure~\ref{fig:SDM}. This means that the gradients of several successive epochs of learning vary greatly. On the other hand, the sensitivity indicators defined above are produced from the gradient. It must be emphasized that other sensitivity indicators, such as `hidden unit efficiency' or the Hessian matrix in the `second order sensitivity analysis', are also strongly influenced by the current set of network parameters (synaptic weights and activation function parameters) and fluctuate significantly in the course of learning.   

To avoid such fluctuations, averaging along the training trajectories was suggested \cite{gorban1990training, gorban1996neural}, implemented and tested \cite{Gorban1995multineuron, gilev1994multineuron}. After averaging over several training epochs, the sensitivity indicators become more robust and do not fluctuate much.

The batch sensitivity indicators for weights tend to zero near the minimizer of the loss function. To avoid this degeneration, small random change of weights and restart of training may be needed. \cite{Tsaregorodtsev_2002} proposed to use the gradient of the norm of the NN output vector instead of the gradient of the loss function.  Systematic testing demonstrated that these batch sensitivity indicators also provide efficient pruning and, at the same time, require much less computational resources.

\section{Pruning procedures and strategies}\label{PrunProc}
The general pruning procedure is shown in Algorithm of NN pruning below. Step of calculating the sensitivity indicators involves several epoches of NN training with the accumulation of the sum of the sensitivity indicators.

{\bf Algorithm of NN pruning.}
\begin{algorithmic}
\STATE Training NN
\LOOP
\STATE Save current NN.
\STATE Calculate sensitivity indicators.
\STATE Select element for modification.
\STATE Modification of NN.
\STATE Retraining modified NN
\IF {total loss function is greater than threshold}
\STATE Break loop
\ENDIF
\ENDLOOP
\STATE Restore the last saved NN
\end{algorithmic}

Procedure for selecting an element for  modification is simple for most of pruning problems: the first candidate is trainable element with minimal value of sensitivity indicator (some of the elements can be marked as non trainable during the pruning process). The only exception is the problem of uniform network simplification. In this case the first candidate for modification is the synaptic weight $w^r_i$ with the smallest sensitivity indicator $\chi^r_i$ such that the $r$ neuron has maximal number of synaptic weights $n^r$. This definition of the first candidate for removal allows us to uniformly simplify the structure of NN.

Modification of NN has different meaning for different pruning problems. For problems involving deleting an element, modification means simple deleting of selected element and possibly related elements. Removing related NN elements involves, for example, removing from a NN a subnetwork whose outputs are not connected (directly or indirectly) to output neurons. For the problem of the reduction of a synaptic weight precision, modification of selected element $w^s_k$ involves replacing it by $v^s_k$ and marking this synaptic weight as non trainable. This means that in any subsequent training periods, this synaptic weight will be constant.

Retraining of the modified NN should begin with the modified network. It is easy to understand that in many cases a modified network can be successfully trained, but training NN with exactly the same structure but with randomly generated initial weights can be unsuccessful.

The described procedure for pruning NN is very simple, but for a rich enough NN this can take a lot of time. There is a very simple modification of the proposed procedure for its acceleration (see Accelerated algorithm of NN pruning below). Let us denote the number of elements for simultaneous modification/deletion as $M$. In the first step $M$ can be half of the total number of elements (input features, neurons, synaptic weights) in NN. After calculating the sensitivity indicators select $M$ elements for modification. If the modified NN cannot be successfully trained, then we return to the last saved NN and divide $M$ by 2: $M=M/2$. Then we repeat the pruning without recalculating the sensitivity indicators. If modification of one element ($M=1$) is unsuccessful, then the last saved NN is the minimal NN, and the procedure stops.

{\bf   Accelerated algorithm of NN pruning.}
\begin{algorithmic}
\STATE Training NN
\STATE $M=$ half of total number of elements to remove.
\LOOP
\STATE Save current NN.
\STATE Calculate sensitivity indicators.
\LOOP
\STATE Select M elements for modification.
\STATE Modification of NN.
\STATE Retraining modified NN
\IF {total loss function is greater than threshold}
\STATE Break internal loop
\ENDIF
\IF {$M>1$}
\STATE $M=M/2$
\ELSE
\STATE Break external loop
\ENDIF
\ENDLOOP
\ENDLOOP
\STATE Restore the last saved NN
\end{algorithmic}

NN after completion of the pruning procedure is {\em minimal}. This means that there are no elements of the considered type which could be modified without destroying NN skills. The notion of a minimal network depends on type of pruning problem to be solved: for feature selection problem the minimal network used a minimal set of input features, for neuron removing problem the minimal NN contains the minimal number of neurons, for synaptic weights precision reduction problem the minimal network contains minimal (usually zero) number of non modified elements. Really it is possible to combine different procedures. For example, we can initially minimise the set of used input features, then minimise the number of neurons, minimise the number of synaptic weights, and, finally, reduce precision of synaptic weights. The final network of this procedure will be minimal from all points of view.

All described pruning procedures require retraining modified NN. In the proposed framework (see Algorithm of NN pruning and Accelerated algorithm of NN pruning), retraining does not require time comparable with initial training \cite{Siegel_2016}.

\section{Knowledge extraction}\label{KnowExtr}

\subsection{Logically transparent NN}

To extract knowledge from trained NN a specially developed pruning procedure \cite{Gorban1999Generation} can be used. The main step in this procedure is to uniformly reduce the number of inputs of each neuron. Interpreting the output of a neuron with one input is a trivial task. Logical interpretation is usually possible for two inputs, it is not so clear for three inputs, and it becomes difficult for a larger number of inputs. On the other hand, if all synaptic weights can be converted to plus or minus one, then interpreting a neuron with three inputs will become simple. 

This means that to extract explicit knowledge from NN, it is necessary to apply a uniform decrease in the number of inputs for each neuron, then the removal of remained redundant synaptic weights can be applied, and finally, all synaptic weights should be converted to one of the values from the set $ S = \{- 1 , 0,1 \} $. 
 Usually, such network allows us to use the step function
\begin{equation}
h(x)=
\begin{cases}
  -1, & \mbox{if } x<0 \\
  1, & \mbox{otherwise}.
\end{cases}
\label{eq:stepFunc}
\end{equation}
instead of the continuous activation functions. 

We call  NN {\em logically transparent} if it is successfully pruned in this way and has no more than three inputs for each neuron.
If at least one of the requirements is not satisfied, then NN is not logically transparent and cannot be represented by a simple ``if-then'' style algorithm. Such NN can be used, for example, to generate ``fuzzy if-then'' rules \cite{Ishibuchi}. Obviously, explicit if-then rules are preferable, but sometimes it is not possible to find such a description. In such situations, an increase in the number of layers and the number of units (neurons) in hidden layers can may help: first, we train an extended NN, and then transform it into the logically transparent form. 

Our definition of logical transparency severely restricts the NN class that we would like to create by pruning. Nevertheless, we cannot expect that the resulting logically transparent NN is unique. The non-uniqueness of knowledge extracted from data is a fundamental property well-known for everybody who works with data-bases AI \cite[Chapter 9]{gorban1998neuroinformatics}. \cite{Breiman_2001} elegantly described this phenomenon as `Rashomon effect' referring to the famous movie  by  Akira Kurosawa. In this film, people retell events, and everyone lies to present themselves better than they really are. 

One comment is needed here. Unlike the situation in the movie Rashomon, the diverse knowledge gained from the data is not a collection of lies. The expected difference between different empirical opinions was well recognised already in ancient philosophy and formulated by Parmenides as two ways, ``The way of truth",  Aletheia, and ``The way of appearance" or opinion, Doxa. The knowledge extracted from the data is an opinion based on appearance, and includes uncertainty that can be attributed to both the data and the extraction procedure. The way of AI based on data  is the way of doxa.

The notions of aletheya and doxa remain in the focus of philosophical discussions \cite{Cordero_2010, Kurfess_2016}. Here, instead of discussion of ancient philosophy, we demonstrate the multiplicity of explicit knowledge produced from data by NN learning and pruning on an example. 

\subsection{Example 1: The semi-empiric political theory of president election in the US}

Consider a database for the problem of predicting the results of the presidential elections in the United States \cite{Lichtman_1981}. Each database entry contains answers to 12 questions:
\begin{enumerate}
  \item Has the incumbent party been in office more than a single term?
  \item Did the incumbent party gain more than 50\% of the vote cast in the previous election?
  \item Was there major third party activity during the election year?
  \item Was there a serious contest for the nomination of the incumbent party candidate?
  \item Was the incumbent party candidate the sitting president?
  \item Was the election year a time of recession or depression?
  \item Was there a growth in the gross national product of more than 2.1\% in the year of the election?
  \item Did the incumbent president initiate major changes in national policy?
  \item Was there major social unrest in the nation during the incumbent administration?
  \item Was the incumbent administration tainted by major scandal?
  \item Is the incumbent party candidate charismatic or a national hero?
  \item Is the challenging party candidate charismatic or a national hero?
\end{enumerate}
This is a binary classification problem with two classes: `P' is the victory of power (incumbent) party and `O' is the victory of opposition (challenging) party. The database contains 31 records (elections from 1860 to 1980). The following NN structure was chosen: two hidden layers with 10 neurons in each and two output neurons, P and O. Network answer was `victory of power party' if the output of neuron P was greater than the output of neuron O, and answer was `victory of opposition party' in the opposite situation. The input was coded as $1$ for the answer `yes' and $-1$ for the answer `no'. The neural network solution for this problem was presented in \cite{gorban1995neural, Waxman}.  \cite{Borisyuk_2005} used a similar questionnaire and  NN for prediction of the result of the UK General Election.

\begin{figure}[ht!]
\centerline{\includegraphics[width=0.7\columnwidth]{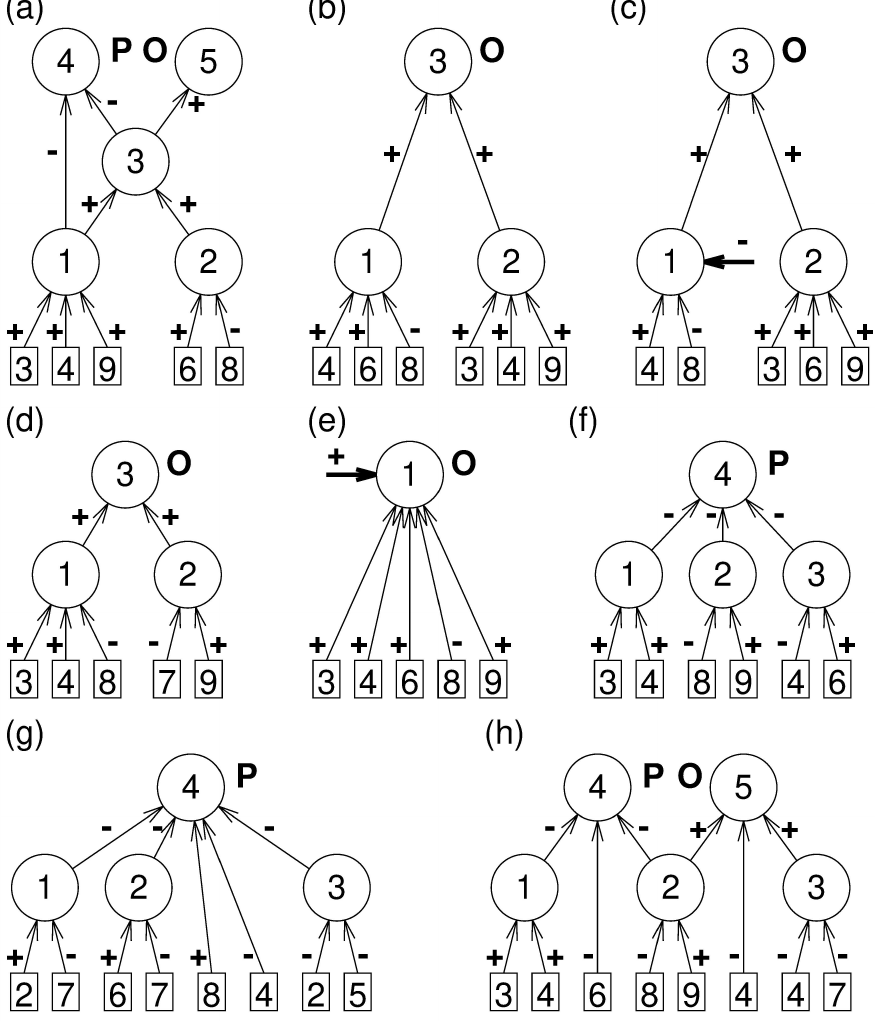}}
\caption{Examples of minimal NNs for the presidential election problem: (a), (b), (c), (d), (f), and (h) are logically transparent, while (e) and (g) are not; a rectangle means input neuron; a bold arrow means a unit bias; a thin arrow means the connection between neurons; the signs ``$+$'' and ``$-$'' mean synaptic weight of 1 or $-1$, respectively; ``O'' (``P'') means the positive signal of neuron corresponds to victory of opposition (power) party}
\label{fig:simplified}
\end{figure}

For the experiment we generated, trained and pruned several networks with different initial weights and different pruning procedures. Minimal NNs are presented in Figure~\ref{fig:simplified}. For networks (a), (b), (c) and (e) the first step in the pruning was input feature selection. As a result these four networks have only five inputs. The second pruning procedure for networks (a), (b), and (c) and the first procedure for networks (d), (f), and (h) was uniform structure simplification (the goal was to have no more than 3 inputs for each neuron). We can see that each neuron of all these NNs has no more than 3 inputs. The next pruning procedure was the removal of neurons. After described pruning procedures, all NNs was pruned by removing of redundant synaptic weights. The final step was to modify the synaptic weights to the values from the set $S=\{-1, 0, 1\}$. In all presented networks, the sigmoid activation function was replaced by the step function \eqref{eq:stepFunc}. Networks (e) and (g) are minimal but not logically transparent because the output neurons of both networks have five inputs. On the other hand, NN in Figure~\ref{fig:simplified}(e) can be described by the logical rule: ``the opposition candidate will win if at least two answers for questions 3, 4, 6, and 9 are 
positive or at least one of these answers is positive and answer for question 8 is negative''. An ordinary person can understand such a rule without much effort.

Let us verbalise the network presented Figure~\ref{fig:simplified}(b). Using medical terminology, the inputs can be called ``symptoms'', the outputs of neurons 1 and 2 are ``syndromes'' and output of neuron 3 is a ``diagnosis''. The first syndrome appears if at least two of the following three symptoms are observed: ``There was a serious contest for the nomination of the incumbent party candidate'', ``The election year was a time of recession or depression'', and ``The incumbent president did not initiate any major changes in national policy'' (note that the negation was used in the last statement). All three statements characterise quality of the current president governance: governance is inadequate if two of these three symptoms are observed. The second syndrome appears if at least two of the following symptoms are observed: ``There was major third party activity during the election year'', ``There was a serious contest for the nomination of the incumbent party candidate'', and ``There was major social unrest in the nation during the incumbent administration''. The second syndrome can be called ``political instability syndrome'': the situation is politically unstable  if at least two of the three symptoms listed above are observed. Neuron 3 produces a positive output if at least one of the syndromes appears.

In all algorithms below we named the syndromes. These names are rather arbitrary and are used only for designations. 
Formulate the algorithm A1 extracted from the NN in Figure~\ref{fig:simplified}(b):
\begin{enumerate}
  \item \emph{Inadequate governance syndrome} appears if at least two of the following symptoms are observed: ``There was a serious contest for the nomination of the incumbent party candidate'', ``The election year was a time of recession or depression'', and ``The incumbent president did not initiate any major changes in national policy''.
  \item \emph{Political instability syndrome} appears if at least two of the following symptoms are observed: ``There was major third party activity during the election year'', ``There was a serious contest for the nomination of the incumbent party candidate'', and ``There was major social unrest in the nation during the incumbent administration''.
  \item The opposite (challenging) party will win if either governance is inadequate or situation is politically unstable.
\end{enumerate}

The NN in Figure~\ref{fig:simplified}(d) represents the following algorithm A2:
\begin{enumerate}
  \item \emph{Syndrome of political instability or stagnation} appears if at least two of the following symptoms are observed: ``There was major third party activity during the election year'', ``There was a serious contest for the nomination of the incumbent party candidate'', and ``The incumbent president did not initiate any major changes in national policy''.
  \item \emph{Syndrome of social instability} appears if both of the following symptoms are observed: ``The a growth in the gross national product was less than 2.1\% in the year of the election'' and ``There was major social unrest in the nation during the incumbent administration''.
  \item The opposite (challenging) party will win if one of instability syndromes appears.
\end{enumerate}

The NN in Figure~\ref{fig:simplified}(a) gives the following algorithm A3:
\begin{enumerate}
  \item \emph{Political instability syndrome} appears if at least two of the following symptoms are observed: ``The election year was a time of recession or depression'', ``There was a serious contest for the nomination of the incumbent party candidate'', and ``There was major social unrest in the nation during the incumbent administration''.
  \item \emph{Inadequate governance syndrome} appears if both of the following symptoms are observed: ``There was a serious contest for the nomination of the incumbent party candidate'' and ``The incumbent president did not initiate any major changes in national policy''.
  \item \emph{Governance ineffectiveness syndrome} appears if at least one of following syndromes is observed: Political instability syndrome or Inadequate governance syndrome.
  \item \emph{The power party potential} is 1 if Political instability syndrome and Governance ineffectiveness syndrome are not observed. Otherwise the power party potential is -1.
  \item \emph{The opposition party potential} is 1 if Governance ineffectiveness syndrome is observed and is -1 otherwise.
  \item The Power(incumbent) party will win if the power party potential is greater than the opposite party potential.
\end{enumerate}

The NN in Figure~\ref{fig:simplified}(c) can be interpreted as the following algorithm A4:
\begin{enumerate}
  \item \emph{Inadequate governance syndrome} appears if both of the following symptoms are observed: ``There was a serious contest for the nomination of the incumbent party candidate'' and ``The incumbent president did not initiate any major changes in national policy''.
  \item \emph{Instability syndrome} appears if at least two of the following symptoms are observed: ``There was major third party activity during the election year'', ``The election year was a time of recession or depression'', and ``There was major social unrest in the nation during the incumbent administration''.
  \item The opposite (challenging) party will win if instability or inadequate governance syndrome appears.
\end{enumerate}

The NN in Figure~\ref{fig:simplified}(e) gives the following algorithm A5:
\begin{enumerate}
  \item The opposite (challenging) party will win if at least two of the following symptoms are observed: ``There was major third party activity during the election year'', ``There was a serious contest for the nomination of the incumbent party candidate'', ``The election year was a time of recession or depression'', ``The incumbent president did not initiate major changes in national policy'', and ``There was major social unrest in the nation during the incumbent administration''.
\end{enumerate}

The NN in Figure~\ref{fig:simplified}(f) gives the following algorithm A6:
\begin{enumerate}
  \item \emph{Syndrome of political instability} appears if at least one of the following symptoms are observed: ``There was major third party activity during the election year'' and ``There was a serious contest for the nomination of the incumbent party candidate''.
  \item \emph{Syndrome of social instability} appears if at least one of the following symptoms are observed: ``The incumbent president did not initiate major changes in national policy'' and ``There was major social unrest in the nation during the incumbent administration''.
  \item \emph{Syndrome of power party consolidation against economic problem} appears if at least one of the following symptoms are observed:
  ``There was no a serious contest for the nomination of the incumbent party candidate'' and ``The election year was a time of recession or depression''.
  \item The opposite (challenging) party will win if at least two of listed syndromes appear.
\end{enumerate}

The NN in Figure~\ref{fig:simplified}(g) can be interpreted as the following algorithm A7:
\begin{enumerate}
  \item \emph{Syndrome low growth after trust} appears if at least one of the following symptoms are observed: ``The incumbent party gained more than 50\% of the vote cast in the previous election'' and ``There was a growth in the gross national product of less than 2.1\% in the year of the election''.
  \item \emph{Syndrome of economic depression} appears if both of the following symptoms are observed: ``The election year was a time of recession or depression'' and ``There was a growth in the gross national product of less than 2.1\% in the year of the election''.
  \item \emph{Syndrome of new and not very trusted president} appears if both of the following symptoms are observed: ``The incumbent party gained less than 50\% of the vote cast in the previous election'' and ``The incumbent party candidate was not the sitting president''.
  \item The opposite (challenging) party will win if at least three of following symptoms and syndromes are observed: Syndrome low growth after trust, Syndrome of economic depression, Syndrome of new and not very trusted president, ``The incumbent president initiated major changes in national policy'', and ``There was no a serious contest for the nomination of the incumbent party candidate''.
\end{enumerate}

The NN in Figure~\ref{fig:simplified}(h) represents the following algorithm A8:
\begin{enumerate}
  \item \emph{Syndrome of power party problem} appears if at least one of the following symptoms are observed: ``There was major third party activity during the election year'' and ``There was a serious contest for the nomination of the incumbent party candidate''.
  \item \emph{Syndrome of weak president} appears if at least one of the following symptoms are observed: ``There was major social unrest in the nation during the incumbent administration'' and ``The incumbent president did not initiate major changes in national policy''.
  \item \emph{Syndrome of consolidation under economic problem} appears if at least both of the following symptoms are observed: ``There was no a serious contest for the nomination of the incumbent party candidate'' and ``There was a growth in the gross national product of less than 2.1\% in the year of the election''.
  \item \emph{The power party potential} is 1 if at least two of the following symptoms and syndromes are not observed: Syndrome of power party problem, Syndrome of weak president and ``The election year was a time of recession or depression''. Otherwise the power party potential is -1.
  \item \emph{The opposite party potential} is 1 if at least two of the following symptoms and syndromes are observed: Syndrome of weak president, Syndrome of consolidation under economic problem, and ``There was a serious contest for the nomination of the incumbent party candidate''. Otherwise the opposite party potential is -1.
  \item The Power (incumbent) party will win if the power party potential is greater than the opposite party potential.
\end{enumerate}

Eight different algorithms were generated. Since all eight algorithms use 8 attributes together, and all attributes are binary, the number of possible different combinations of these attributes is $2^8=256$.

In 113 cases (more than 44\%), the answers of all eight algorithms are the same. There are 47 cases where one algorithm contradicts to seven other algorithms, 36 cases where two algorithms are disagree with the other six algorithms. There are 45 cases where three algorithms are in opposition to other five algorithms. And, finally, there are 15 cases where voting of algorithms are divided equally. How we can use such a ``pluralism'' in various real life situations? First of all, we can try to collect data at those points where there is no consensus. Such new points can be used to retrain and identify really correct algorithms for extended database. Another usage of this collection of eight algorithms can be a ensemble of experts with simple voting. In this case, our multiple experts can predict the answers in the new points with some ``probability' or, better to say, ``certainty".

\subsection{Example 2: Pruning of inputs for recognition of handwritten digits (MNIST data)}

We compared efficiency of pruning with zero-order sensitivity indicators and with our first order indicators on the MNIST database of handwritten digits. Used data set contains 10,000 grayscale images  of handwritten digits (1,000 images of each digit) with resolution $28\times 28$ pixels. This is a subset of  MNIST dataset \url{http://yann.lecun.com/exdb/mnist/}. 55 input signals were removed because they were constant in the dataset.   
Used neural network contains 784 input signals with min-max normalisation into [-1,1] interval, 10 neurons in hidden layers with sigmoid activation function 
$$\phi(x)=\frac{2}{1-\exp(-2x)} -1.$$

Each class  was randomly split into training (70\%), test (15\%), and validation (15\%) sets. The training set was used to train the networks and  to define candidates for removing. The validation set is used to estimate appropriateness of removing selected weights. The test set was used to estimate the final pruning quality. 

\begin{figure} 
\centerline{\includegraphics[width=0.7\columnwidth]{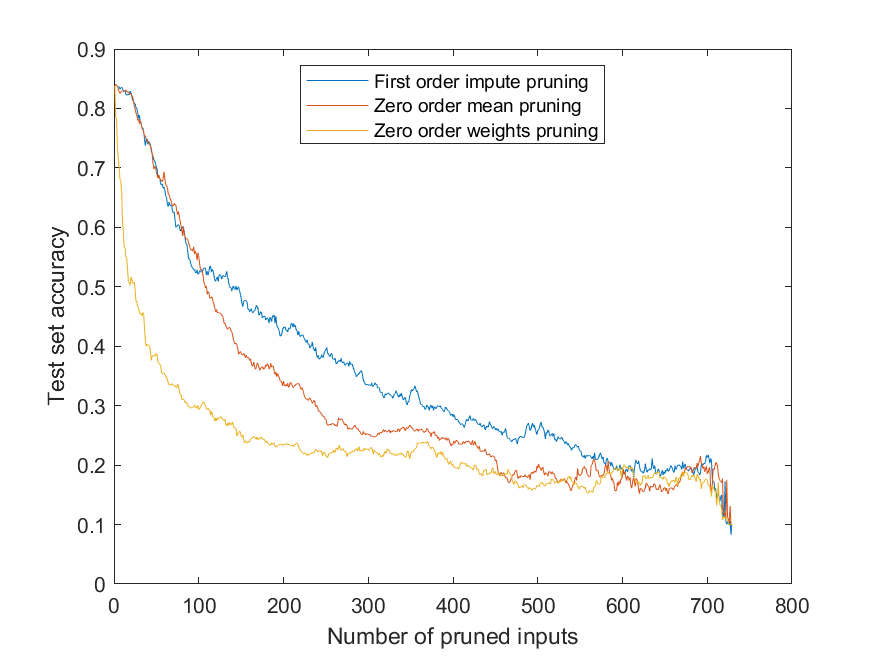}}
\caption{Efficiency of three pruning strategies. The accuracy of pruned networks is presented as a function of the number of synapses removed as a result of pruning.}
\label{fig:MNIST}
\end{figure}

\section{Pruning as sparse linear regression }\label{sparse}

The pruning problem in its simplest version (reduce the number of synapses) is closely related to the well-elaborated sparse linear  regression problem. Sparse regression is needed for feature selection in statistical learning when the number of predictor variables is too large. Traditional regression methods can fail because insufficient number  of empirical observations, hidden correlations between predictors, and other difficulties \cite{Friedman_2012,Bertsimas_2020}. 

The sparse linear regression problem is formulated as follows \cite{Friedman_2012}. Let $N$ observation be given:
\begin{equation}
\label{DataRegres}
(y_i, \boldsymbol{x}_i), \; i=1,\ldots N.
\end{equation}
Here, $y$ is the outcome, and $\boldsymbol{x}=(x_1, \ldots , x_n)$ is the $n$-dimensional input vector. The values of $x_i$ may be original inputs or some functions constructed from them. The goal is to find the regression function
$$\hat{y}(\boldsymbol{x})=w_0+\sum_{j=1}^n w_j x_j$$
that estimates $y$.

The solution should minimize the expectation of risk function. It is evaluated on the empirical distribution (\ref{DataRegres}). If the loss function for one example is $l(y_i,\hat{y}(\boldsymbol{x}_i))$ then the empirical risk function is defined as
$$\hat{R(\boldsymbol{w})}=\frac{1}{N}\sum_{i=1}^N l(y_i,\hat{y}(\boldsymbol{x}_i)).$$
The popular loss functions are: the squared error, $(y-\hat{y})^2$ and the squared error with margin $\varepsilon$,  $l(y,\hat{y})=(|y-\hat{y}|-\varepsilon)^2 H(|y-\hat{y}|-\varepsilon)$, where $H$ is the Heaviside unit step function. 

Minimization of $\hat{R}(\boldsymbol{w})$ can give a poor estimate of the vector of coefficients, especially when the number of observations is relatively small. Therefore, many regularization additions were invented. The coefficients are defined as
\begin{equation}\label{Regul}
\hat{\boldsymbol{w}}=\arg \min_{\boldsymbol{w}} \left[\hat{R(\boldsymbol{w})}+ \lambda P(\boldsymbol{w})\right],
\end{equation}
where $P(\boldsymbol{w})$ is the penalty function and the coefficient $\lambda>0$ regulates the penalty strength.

The popular penalty functions are  the power penalties,
$$P_{\gamma}(\boldsymbol{w})=\sum_{j=1}^n |w_j|^{\gamma}, \; \gamma>0.$$
For $\gamma\leq 1$ the minimization (\ref{Regul}) can give  the sparse solution, where  $w_j=0$ for some $j$. The value of $\lambda$ regulates the number of zero coefficients estimates. It increases with $\lambda$. Smaller values of $\gamma$ may provide better sparsification but for $\gamma<1$ the minimizing functional in (\ref{Regul}) becomes non-convex. Nevertheless, even for this case, an efficient piecewise-quadratic approximation of loss function gives a computationally efficient solution \cite{GorbanPQSQ_2016}. An application-oriented review of modern popular sparse regression methods was presented by \cite{Bertsimas_2020}.    

Any method of sparse linear regression can be applied to pruning of trained neural network. For a given set of examples, presented by the vectors of the network input signals, the network produces all the intermediate signals. As a result, for each neuron we have the set of input signals and the corresponding outputs of the linear combiner  (\ref{combiner})). This is a linear regression problem and we can find the optimal set of weights values to solve this problem with the required accuracy and the maximal sparsity.

The sparse linear regression approach to pruning decreases the number of synapses with preservation of accuracy. It is local by its nature (works with different neurons separately) and does not require multiple learning iteration. At the same time, flexible formation of logically transparent network may not be possible by such a local method.     

Applying sparse linear regression for pruning requires an estimate of the required accuracy for each neuron. The accuracy requirement for the whole network should be decomposed into requirements for each single neuron. ``Backpropagation of accuracy" calculates the maximum permissible errors possible for the signals and parameters of each network element, from the condition that the vector of the output signals of the network must be calculated with a given accuracy \cite{Senashova_1997}.

\section{Discussion}\label{discus}

In this paper, we   presented a general description of the NN pruning problems and reviewed the solutions based on the first order sensitivity indicators. This approach was started 30 years ago and has undergone a number of rediscoveries and modifications.
The sensitivity analysis was developed not only for synaptic weights, but also for any signals or group of signals in NN. This approach allows us to evaluate sensitivity to each neuron  directly. 

Pruning based on the first order sensitivity indicators can be easily applied to CNN. The main advantage of CNN is uniformity of all calculations. This means that procedure for removing of one synaptic weight becomes useless: it can complicate calculations instead of simplifying them. Essentially more reasonable is operation of removing of filters \cite{Han_2015} or, even, channels \cite{He_2017}. Algorithms of sensitivity indicators estimation \cite{gorban1990training, gorban1996neural, mirkes1998neurocomputer} calculate them for each filter and each channel. This allows us to apply effectively  the Algorithm of NN pruning to CNN. Procedures of  precision reduction can be applied directly to DL NN.

The presented pruning procedures can be implemented in a unified framework: all the  procedures correspond to the same Accelerated algorithm of NN pruning and differ in the assessment of sensitivity indicator and the selection of candidates for modification. There are two different ways to evaluate the sensitivity indicator: the sensitivity to synaptic weight or the sensitivity to signals (input, output, or intermediate). There are also two described approaches for selecting candidates for modification: globally or in several local sets (for the uniform simplification of all neurons or other structural blocks).

The special sequences of pruning procedures allow us to form a special class of networks: logically transparent NNs, and then create a verbal description of an explicit algorithm for solving the problem. 
This approach was demonstrated on the problem of forecasting the result of the USA Presidential Election and eight explicit algorithms was generated. 

How this knowledge can be used? We do not like to discuss the `real politics' here. Imagine  instead a strategic game `Presidential Elections'. Apply algorithm A1 (Figure~\ref{fig:simplified}(b)). The opposition party should make one of the syndromes true to win the election: (1) the syndrome of inadequate governance or (2) the syndrome of political instability. For for both syndromes the symptom (4) is important: 
``There was a serious contest for the nomination of the incumbent party candidate''. Therefore, it seems to be reasonable to invest much efforts into generation or demonstration to general public the contest and conflicts in power party. Then, there are three direction of investments: (i) active propaganda that the president has not done anything new, and all changes in national politics are just cosmetics (this is important - not wrong changes, but no significant changes at all), (ii) investments into activity of a third party, and (iii) investments into social protest. The opposition party must choose the main direction (not all three are necessary) and properly allocate resources. The counterplay of the presidential party is also clear. This is followed by a reflexive game of intelligence and analysis of the opponent's strategy from both sides. Of course, the real life can be more complicated, this analysis may be too superficial, but instructive  nonetheless. 

Proposed method usually gives several explicit algorithms. Is this an advantage or disadvantage? We find this property useful and reasonable. To further improve the theory and discriminate models  we can identify areas where different algorithms will produce different results, and collect new data from these areas. New data can be used to reduce the number of models or falsify all of them and create new explicit algorithms.

Each reduced network (and the corresponding algorithm) can be considered as an intellectual agent and the set of algorithm is a set of intelligent agents for problem solving. These agents can vote and create a committee decision \cite{Hansen_1990, Gilev_1991, Jiang_2014}.

We are sure that the technology that gives the only variant of explicit knowledge is unreliable, and the non-uniqueness of the result is a fundamental property of the production of explicit knowledge from data. At the roots of sciences we can find Parmenides dividing the path of knowledge into the path of truth (Aletheya) and the path of opinion (Doxa) \cite{Mourelatos_2008}. Doxa is non-unique and variable. The knowledge based on data definitely belong to the way of Doxa, whereas theoretical science used the idea of undoubted truth  (Aletheya) as a unattainable ideal. Data-driven AI definitely belongs to the Doxa path and produces multiple empirical laws (opinions). At the same time, it demonstrates that this path can also be beautiful and, as we expect, can compete with the Alethea path.   

It seems to be a non-trivial task to find the physiologically relevant pruning mechanism  because the back-propagation procedures and gradient learning algorithms are not found in the brain.  Nevertheless, the mechanism hidden in the first order sensitivity indicators may have the physiological representation if we properly reformulate this approach avoiding back-propagation terms. The main principle is: the pruning criteria are connected with the learning rate. {\it What controls training should control pruning.} 

Another important idea is locality. We expect that the dynamics of pruning, as well as the difference in plasticity of different synapses, are driven by local processes in the brain. ``Optimizing global network architecture using local synaptic rules'' may be considered and the main problem of biologically relevant pruning \cite{Scholl_2020}.

Synapses are a costly resource whose efficient utilization is a major optimization goal in brain development. This biological pruning is assumed to be closely connected with the learning process \cite{Chechik_1998}.
According to the Hebbian rule and its various generalizations \cite{Oja_1998, Friston_1993, Tyukin_2019}, the learning rate of the synapse  is controlled by the average products of activity of the neurons connected by them (with some saturation effects). Taking into account the correlated activity of the neurons that are connected by a short path, we can formulate a general (but less formal) rule: learning is controlled  by the positive functional of activity of the neurons in a vicinity of the element in the connectome. The vicinity in the connectome could differ significantly from just a geometrially close neurons. The second mechanism is caused by the neuronal--glial interactions. High local activity of neurons activate the close glial cells: microglia, astrocytes, and other cells \cite{Jakel2017} and their feedback regulates the synapse plasticity. 

For example, astrocytes interact with neurons via chemicals diffused
in the extracellular space. Calcium elevations occur in response to the increased
concentration of the neurotransmitter released by spiking neurons when a group of them
fire coherently. For example, coherent spiking of a group of local neurons stimulate connected with them astrocytes and calcium elevation,  gliotransmitters are released by activated astrocytes and modulate the synaptic connections. This modulation lasts from a dozen seconds to a dozen minutes and contributes to both short- and long-term synaptic plasticity. These processes control selection of the ``most important'' synapses and may be the key elements in emergence of working memory  \cite{Gordleeva_2021}. These ideas and mechanisms are promising, but a biologically adequate theory of selection of the most important connections has yet to be developed. 
 
\section{Conclusion: The future of XAI} 
 
Now, after several years of XAI program and many years of our pruning research we can formulate several general predictions:

\begin{itemize}
\item The explicit rules extracted from the neural network or other AI systems are significantly  non-unique. This is a fundamental restriction. We can state that this restriction is in the core of the semi-empirical knowledge. We never receive a unique truth. And if we add some additional restrictions for the selection of unique results, it will be abuse against nature.
\item The multiplicity of truth should increase fast with the data dimensionality. Therefore, the natural area of XAI applications will be in low- and medium-dimensional problems. 
\item Localization of reasoning, which means choosing individual rules for each sample or a small cluster of samples, reduces the local number of rules for each particular case, but does not provide a solution to the whole problem.
\item Are we and our users ready to work with this enormous collections of rules? We cannot answer this question a priory. We can guess that there may appear a new profession ``XAI Engineer'', who will work with large sets of local and global rules with special additional software instruments. Regular AI users cannot do this job outside of small and attractive low-dimensional examples. 
 \end{itemize}

\section*{Conflict of Interest Statement}

The authors declare that the research was conducted in the absence of any commercial or financial relationships that could be construed as a potential conflict of interest.

\section*{Author Contributions}
ANG theory development, EMM software development and numerical experiments, ANG and EMM writing the text and edition

\section*{Funding}
The project is supported by the Ministry of Science and Higher Education of the Russian Federation (Project No 075-15-2020-808).

\section*{Acknowledgments}
The content of this manuscript has been presented in part at the International Joint Conference on Neural Networks (IJCNN), 2020 \cite{Mirkes_2020}.


\section*{Data Availability Statement}
The datasets analysed for this study can be found in Table 2 of \cite{Lichtman_1981}.


\bibliographystyle{plain}





\end{document}